\documentclass[accepted]{uai2024} 
                
\usepackage[american]{babel}

\usepackage{natbib} 
\usepackage{mathtools} 
\usepackage{booktabs} 
\usepackage{tikz} 

\usepackage{algorithm}
\usepackage{algorithmic}

\usepackage{amssymb} 
\usepackage{natbib}
\usepackage{multirow}
\usepackage{diagbox}
\usepackage{nccmath}
\usepackage{ascmac}
\usepackage{here}

\usepackage{xcolor}

\usepackage{booktabs}
\usepackage{url}
\newcommand \indep{\mathop{\perp\!\!\!\perp}}

\DeclareMathOperator{\E}{\mathbb{E}}

\DeclareMathOperator{\R}{\mathbb{R}}
\DeclareMathOperator{\pr}{\mathrm{P}}
\DeclareMathOperator{\I}{\mathbf{I}}

\DeclareMathOperator{\F}{\mathrm{F}}

\usepackage{bm}

\DeclareMathOperator{\birho}{\bm{\rho}}

\newcommand{\bic}{\textbf{\textit{c}}}

\newcommand{\biX}{\textbf{\textit{X}}}
\newcommand{\bix}{\textbf{\textit{x}}}
\newcommand{\biw}{\textbf{\textit{w}}}
\newcommand{\biW}{\textbf{W}}
\newcommand{\bir}{\textbf{\textit{r}}}

\newcommand{\Yz}{Y^{0}}
\newcommand{\Yo}{Y^{1}}

\newcommand{\nf}{d}

\usepackage{txfonts}

\usepackage{cleveref}

\definecolor{navyblue}{rgb}{0.0, 0.0, 0.5}
\hypersetup{
    colorlinks=true,
    allcolors=navyblue
}

\title{Differentiable Pareto-Smoothed Weighting \\
for High-Dimensional Heterogeneous Treatment Effect Estimation}

\author[1]{Yoichi Chikahara}
\author[2*]{Kansei Ushiyama}
\affil[1]{%
    NTT Communication Science Laboratories,
    Kyoto, Japan
}
\affil[2]{%
    The University of Tokyo,
    Tokyo, Japan
}

\begin{document}
\maketitle

\begin{abstract}
  There is a growing interest in 
  estimating heterogeneous treatment effects across individuals
  using their high-dimensional feature attributes. 
  Achieving high performance in
  such high-dimensional heterogeneous treatment effect estimation 
  is challenging
  because in this setup, it is usual that 
  some features induce sample selection bias
  while others do not but are predictive of potential outcomes.
  To avoid losing such predictive feature information,
  existing methods learn separate feature representations
  using inverse probability weighting (IPW).
  However, 
  due to their numerically unstable IPW weights,
  these methods suffer from estimation bias under a finite sample setup.
  To develop a numerically robust estimator by weighted representation learning,
  we propose a differentiable Pareto-smoothed weighting framework
  that replaces extreme weight values in an end-to-end fashion.
  Our experimental results show that by effectively correcting the weight values, our proposed method outperforms the existing ones, 
  including traditional weighting schemes.
  Our code is available at \url{https://github.com/ychika/DPSW}.
\end{abstract}

\renewcommand{\thefootnote}{\fnsymbol{footnote}}
\footnotetext[1]{Work during the summer internship at NTT Communication Science Laboratories.}
\renewcommand{\thefootnote}{\arabic{footnote}}
\setcounter{footnote}{0}

\section{Introduction} \label{sec-intro}

In this paper, we tackle the problem of estimating heterogeneous treatment effects across individuals from high-dimensional observational data. 
This problem, which we call 
high-dimensional heterogeneous treatment effect estimation,
offers the following crucial applications:
the evaluation of medical treatment effects 
from numerous attributes \citep{shalit2020can}
and the assessment of the advertising effects from each user's many attributes 
\citep{bottou2013counterfactual}.

One fundamental difficulty
in high-dimensional heterogeneous treatment effect estimation 
is the sample selection bias induced by \textit{confounders}, 
i.e., the features of an individual that 
affect their treatment choices and outcomes.
For instance, in the case of medical treatment, 
age is a possible confounder:
Elderly patients avoid choosing surgery due to its risk
and generally suffer from higher mortality \citep{zeng2022uncovering}.
Due to their treatment choice imbalance, 
there are often fewer records of elderly patients 
who have received surgical treatments.
As a result, accurately predicting surgical outcomes is difficult in this age cohort,
thus complicating heterogeneous treatment effect estimation.

To address such sample selection bias,
it is crucial to determine how to break the dependence 
of treatment choice on confounders.
A promising approach for a high-dimensional setup is
representation learning \citep{shalit2017estimating},
which estimates the potential outcomes under different treatment assignments
by extracting the \textit{balanced} feature representation
that is learned 
such that its distribution is identical 
between treated and untreated individuals.

However, since this approach converts all input features to 
a single balanced representation,
if some features are \textit{adjustment variables} 
(a.k.a., \textit{risk factors}) \citep{brookhart2006variable},
which are unrelated to sample selection bias 
but useful for outcome prediction,
such useful feature information may be inadvertently eliminated,
leading to inaccurate treatment effect estimation \citep{sauer2013review}.
This issue is serious, especially in high-dimensional setups
where input features often contain adjustment variables.
Moreover,
due to the lack of prior knowledge about the features,
it is difficult for practitioners
to correctly separate adjustment variables from confounders.
Such feature separation might be impossible
if one attempts to input the feature embeddings 
of a complex object (e.g., texts, images, and graphs)
that are constructed from pre-trained generative models,
including large language models (LLMs) \citep{keith2020text}.

To fulfill such important but often overlooked needs,
several data-driven feature separation methods have been proposed
\citep{kuang2017treatment,hassanpour2020learning,kuang2020data,wang2023treatment}.
Among them,
\textit{disentangled representations for counterfactual regression} (DRCFR) \citep{hassanpour2020learning}
aims to avoid losing useful feature information 
for heterogeneous treatment effect estimation 
by learning different representations for confounders and adjustment variables.
To achieve this,
this method employs a technique of inverse probability weighting (IPW) \citep{rosenbaum1983central},
which performs weighting based on the inverse of the probability called a \textit{propensity score}.
However, 
such weights often take extreme values 
(especially in high-dimensional setups \citep{li2017matching}),
and even slight propensity score estimation error may lead to
large treatment effect estimation error.

To resolve this issue, 
we develop a weight correction framework 
that utilizes a technique in extreme value statistics,
called \textit{Pareto smoothing} \citep{vehtari2024pareto},
which replaces the extreme weight values with 
the quantiles of generalized Pareto distribution (GPD).
Indeed, \citet{zhu2020pareto} already adopted Pareto smoothing
and empirically showed that 
it can construct a numerically stable estimator 
of the average treatment effect (ATE) over all individuals.

To estimate heterogeneous treatment effects across individuals,
we propose a Pareto-smoothed weighting framework that 
can be combined with the weighted representation learning approach.
However, achieving this goal is difficult 
because weight correction with Pareto smoothing requires
the computation of the rank (position) of each weight value;
this computation is non-differentiable and prevents gradient backpropagation.
To overcome this difficulty,
we utilize a \textit{differentiable ranking} technique
\citep{blondel2020fast}
and establish a differentiable weight correction framework 
founded on Pareto smoothing.
This idea of combining Pareto smoothing and differentiable ranking,
both of which have been studied in completely different fields 
(i.e., extreme value statistics and differentiable programming),
allows for effective end-to-end learning for high-dimensional heterogeneous treatment effect estimation.

\textbf{Our contributions} are summarized as follows.
\begin{itemize}
    \item We propose a differentiable Pareto-smoothed weighting framework 
    that replaces extreme IPW weight values in an end-to-end fashion. 
    To make this weight replacement procedure differentiable,
    we utilize a \textit{differentiable ranking} technique
    \citep{blondel2020fast}.
    \item Exploiting differentiability,
    we build our weight correction framework on 
    the neural-network-based weighted representation learning method 
    (i.e., DR-CFR \citep{hassanpour2020learning})
    and perform data-driven feature separation
    for high-dimensional heterogeneous treatment effect estimation.
    \item We experimentally show that
    our method effectively learns the feature representations
    and outperforms the existing ones, 
    including traditional weighting schemes.
\end{itemize}

\section{Preliminaries}

\subsection{Problem Setup} \label{sec-problem}

Suppose we have a sample of $n$ individuals, $\mathcal{D} = \{(a_i, \bix_i, y_i)\}_{i=1}^n \overset{i.i.d.}{\sim} \pr(A, \biX, Y)$, 
where $A \in \{0, 1\}$ is a binary treatment ($A=1$ if an individual is treated; otherwise $A=0$), 
$\biX = [X_1, \dots, X_{\nf}]^{\top}$ is the $\nf$-dimensional features (a.k.a., covariates), and $Y$ is an outcome. 
Let $\Yz$ and $\Yo$ be \textit{potential outcomes}  
i.e., the outcomes when untreated ($A=0$) and when treated ($A=1$),
which are given by $Y = A \Yo + (1 - A) \Yz$. 
A treatment effect for an individual is defined as their differences, i.e.,  $\Yo - \Yz$
 \citep{rubin1974estimating}.

We consider the heterogeneous treatment effect estimation problem,
where we take (as input) sample $\mathcal{D}$ 
and feature values $\bix$
and output the estimate of 
 a conditional average treatment effect (CATE) conditioned on $\biX = \bix$:
\begin{align}
    \mathbb{E}\left[\Yo - \Yz \mid \biX = \bix\right]. \label{eq-CATE}
\end{align}
CATE is an average treatment effect in a subgroup of individuals who have identical feature attributes $\biX = \bix$. 

To estimate the CATE in \eqref{eq-CATE}, we make two standard assumptions. 
One is \textit{conditional ignorability}, $\{\Yz, \Yo\} \indep A \mid \biX$;
this conditional independence is satisfied 
if features $\biX$ contain all of the confounders 
and include only \textit{pretreatment variables},
which are not affected by treatment $A$.
The other is \textit{positivity}, $0 < \pi(\bix) < 1$ for all $\bix$, where $\pi(\bix) \coloneqq \pr(A=1 \mid \biX = \bix)$ is 
a conditional distribution model called a propensity score.

Our goal is to achieve a high CATE estimation performance in 
a high-dimensional setup, 
where the number of features $d$ is relatively large.
Under this setup,
removing the sample selection bias
by transforming all features $\biX$ into a single balanced representation 
might be overly severe,
which leads to inaccurate treatment effect estimation.
Although several representation learning methods 
aim to avoid such an overly severe balancing 
\citep{kuang2017treatment,wang2023treatment}, 
most are designed to estimate ATE, not CATE.
By contrast,
the DRCFR method \citep{hassanpour2020learning} 
focuses on CATE estimation
and effectively balances high-dimensional features $\biX$
by weighted representation learning.

\subsection{Weighted Representation Learning} \label{sec-drcfr}

\begin{figure}[t]
    \centering
    \includegraphics[width=0.35\textwidth]{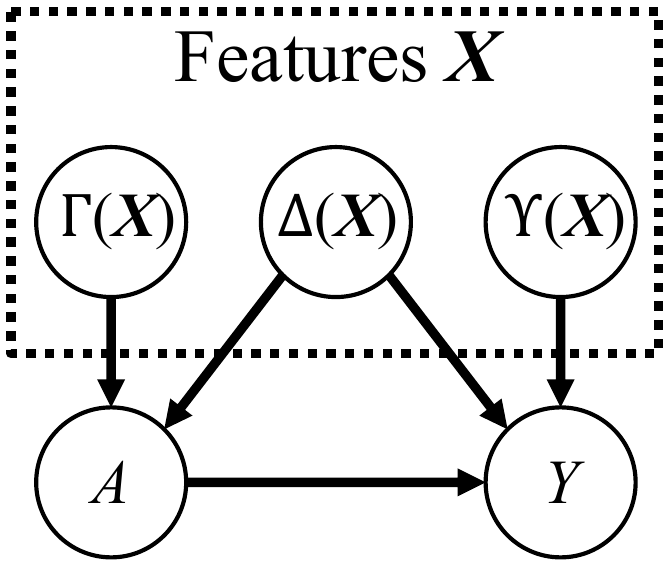}
    \caption{Graphical model illustration of DRCFR method}
    \label{fig1}
\end{figure}

DRCFR \citep{hassanpour2020learning} is based on 
the graphical model in \Cref{fig1}, where 
treatment $A$ is determined 
by functions $\Gamma(\biX)$ and $\Delta(\biX)$,
while outcome $Y$ is given by $\Delta(\biX)$ and $\Upsilon(\biX)$.
Following this model,
$\Gamma(\biX)$, $\Delta(\biX)$, and $\Upsilon(\biX)$ are formulated
as three neural network encoders,
each of which extracts the representation of 
\textit{instrumental variables}
(i.e., the features that influence
$A$ but not $Y$),
confounders,
and adjustment variables, respectively.
For example, in medical treatment setups, 
a possible instrumental variable, confounder, and adjustment variable
might be income, age, and smoking habits, respectively.

To learn the representations of such features,
DRCFR minimizes the weighted loss
by computing the weights 
based on the inverse of propensity scores,
given by $\pi(\Gamma(\biX), \Delta(\biX))$.

A strong advantage of DRCFR is that 
it can avoid losing the information of adjustment variables,
represented by $\Upsilon(\biX)$,
which is useful for outcome prediction.
However, 
since the inverse of conditional probability, $\pi(\Gamma(\biX), \Delta(\biX))$,
often yields extreme values,
under finite sample settings,
even a slight estimation error of $\pi$ 
leads to a large weight estimation error.
This numerical instability of weight estimation 
makes it difficult to achieve high CATE estimation performance.

\section{Proposed Method}

To improve the estimation stability of weighted representation learning,
we propose a differentiable weight correction framework 
that can be used in an end-to-end fashion.

\subsection{Overview} \label{sec-obj}

To estimate the CATE in \eqref{eq-CATE},
following DRCFR \citep{hassanpour2020learning},
we perform weighted representation learning.
We learn three model components:
the feature representations 
(i.e., $\Gamma(\biX)$, $\Delta(\biX)$, and $\Upsilon(\biX)$ in \Cref{fig1}),
propensity score model $\pi(\Gamma(\biX), \Delta(\biX))$,
and outcome prediction models 
$h^0(\Delta(\biX), \Upsilon(\biX))$ and $h^1(\Delta(\biX), \Upsilon(\biX))$,
where $h^0$ and $h^1$ are used 
to predict potential outcomes $\Yz$ and $\Yo$.

DRCFR jointly optimizes these three model components 
by minimizing the weighted loss.
However, we empirically observed that such a joint optimization is difficult.
A possible reason is that
the loss function dramatically changes with the IPW weights 
and hence substantially varies with the parameter values of propensity score $\pi$.
For this reason, 
we separately learn $\pi$ 
and perform an alternate optimization 
that repeatedly takes the following two steps.

First, we learn propensity score $\pi$ (while fixing the other model parameters)
by minimizing the cross entropy loss:
\begin{align}
    \underset{\pi}{\text{min}}\ - &\frac{1}{n}  \sum_{i=1}^n  
    \Bigl( a_i \log(\pi(\Gamma(\bix_i), \Delta(\bix_i))) \nonumber \\
        &+  (1 - a_i) \log(1 - \pi(\Gamma(\bix_i), \Delta(\bix_i))) \Bigr) 
    + \lambda_{\pi} \Omega(\pi), \label{eq-obj-prop}
\end{align}
where $\Omega(\cdot)$ is a regularizer that penalizes the model complexity, 
and $\lambda_{\pi} > 0$ is a regularization parameter.

Second, we learn the other model parameters (with $\pi$'s parameters fixed)
by minimizing the weighted loss: 
\begin{align}
    \underset{\Gamma, \Delta, \Upsilon, h^0, h^1}{\text{min}}\ &\frac{1}{n} 
    \sum_{i=1}^n w_i l(y_i, h^{a_i}(\Delta(\bix_i), \Upsilon(\bix_i)))  \nonumber \\
    &+ \lambda_{\Upsilon} \mathrm{MMD}\left(\bigl\{\Upsilon(\bix_i)\bigr\}_{i: a_i=0}, \bigl\{\Upsilon(\bix_i)\bigr\}_{i: a_i=1}\right)  \nonumber \\
    &+ \lambda_{-\pi} \Omega\left(\Gamma, \Delta, \Upsilon, h^0, h^1\right), \label{eq-obj-other}
\end{align}
where $w_i$ is the weight that is given by propensity score $\pi(\Gamma(\bix_i), \Delta(\bix_i))$,
$l$ is the prediction loss for outcome $y_i$, 
$\lambda_{\Upsilon} > 0$ and $\lambda_{-\pi} > 0$ are regularization parameters,\footnote{$-\pi$ denotes the other model components than $\pi$.}
and $\mathrm{MMD}$ denotes the kernel maximum mean discrepancy (MMD) \citep{gretton2012kernel}, 
which measures the discrepancy between empirical conditional distributions
 $\hat{\pr}(\Upsilon(\biX) \mid A=0)$ and $\hat{\pr}(\Upsilon(\biX) \mid A = 1)$.
Regularizing this MMD term prohibits $\Upsilon$ from having any information about treatment $A$,
thus making $\Upsilon(\biX)$ 
a good representation of the adjustment variables.

To achieve a high CATE estimation performance, 
how to compute the weight value (i.e., $w_i$ in \eqref{eq-obj-other})
is essential.
In the DRCFR method \citep{hassanpour2020learning},
weight is formulated using \textit{importance sampling},
which employs a density-ratio-based weight 
to construct a weighted estimator of expected value. 
To estimate the expected outcome prediction losses 
over the observed individuals ($A=a_i$) 
and the unobserved individuals ($A=1-a_i$),
DRCFR formulates the weight as the sum of two density ratios:
\begin{align}
w_i &= \frac{\pr(\Gamma(\bix_i), \Delta(\bix_i) \mid A=a_i)}{\pr(\Gamma(\bix_i), \Delta(\bix_i) \mid A=a_i)} + \frac{\pr(\Gamma(\bix_i), \Delta(\bix_i) \mid A=1 - a_i)}{\pr(\Gamma(\bix_i), \Delta(\bix_i) \mid A=a_i)} \nonumber \\
&= 1 + \frac{\pr(A=1 - a_i \mid \Gamma(\bix_i), \Delta(\bix_i)) }{\pr(A=1 - a_i)} \frac{\pr(A=a_i)}{\pr(A=a_i | \Gamma(\bix_i), \Delta(\bix_i)) } \nonumber \\
&= 1 + \frac{\pr(A=a_i)}{\pr(A=1-a_i)} \left( \frac{1}{\pr(A=a_i \mid \Gamma(\bix_i), \Delta(\bix_i))} - 1\right)  \label{eq-IPW} \\
&\propto \frac{1}{\pr(A=a_i \mid \Gamma(\bix_i), \Delta(\bix_i))}  \coloneqq \frac{1}{\pi_{a_i}(\Gamma(\bix_i), \Delta(\bix_i))} \nonumber 
\nonumber, 
\end{align}
where $\pi_{a_i}(\biX) = a_i \pi(\biX) + (1 - a_i) (1 - \pi(\biX))$.
Here weight $w_i$ is 
proportional to the inverse of propensity score 
$\pi_{a_i}(\Gamma(\bix_i), \Delta(\bix_i))$.
Since such an IPW weight often takes an extreme value,
the weight estimation is numerically unstable,
leading to inaccurate CATE estimation.
This issue is serious in a high-dimensional setup
due to the difficulty of correctly estimating propensity scores
\citep{assaad2021counterfactual}.

To resolve this issue, 
we improve the weight stability 
by replacing an extreme value of $w_i$ in Eq. \eqref{eq-IPW}
with a weight stabilization technique,
called Pareto smoothing.

\subsection{Weight Correction via Pareto Smoothing}

Pareto smoothing \citep{vehtari2024pareto} is a technique 
for improving the weight stability of importance sampling.

According to \citet{vehtari2024pareto},
this technique has two advantages.
First, it can yield a less biased estimator,
compared with weight truncation, which replaces extreme weights 
naively with constants 
\citep{crump2009dealing,ionides2008truncated}:
\begin{align}
    w^{\mbox{Trunc.}}_i \coloneqq \left\{
        \begin{array}{ll}
            L & \mathrm{if}\quad w_i < L \\
            w_i & \mathrm{if}\quad  L \leq w_i < U \\
            U & \mathrm{if}\quad U \leq w_i \\
        \end{array}
    \right., \label{eq-Trunc}
\end{align}
where $L > 0$ and $U > 0$ are the truncation thresholds.
Second, it can be combined with self-normalization, 
which prevents the weights from being too small or too large relative to each other
by dividing each weight value by 
its empirical mean under identical treatment assignment:
\begin{align}
    w^{\mbox{Norm.}}_i \coloneqq 
    \frac{w_i}{\overline{w^{A=a_i}}},\ \mbox{where}\
    \overline{w^{A=a_i}} = \frac{\sum_{j=1}^n  \I(a_j=a_i) w_j}{\sum_{j=1}^n \I(a_j=a_i)}.
        \label{eq-Norm}            
\end{align}
Here $\I(a_j = a_i)$ is an indicator function 
that takes $1$ if $a_j = a_i$; otherwise, 0. 
We experimentally confirmed that performing self-normalization over Pareto-smoothed weights leads to better CATE estimation performance (\Cref{sec-semiexp}).

To construct a weighted estimator 
that is numerically robust to weight estimation error,
Pareto smoothing 
replaces the extremely large weight values 
with GPD quantiles in two steps:
GPD parameter estimation and weight replacement.

\subsubsection{GPD Parameter Fitting} \label{sec-fit}

First, we fit the GPD parameters to large IPW weight values.

Suppose that random variable $W$ follows the GPD. 
Then 
its GPD cumulative distribution function is defined as 
\begin{align}
    \F(w)= \left\{ \begin{array}{ll}
      1 - \left(1 + \frac{\xi(w - \mu)}{\sigma}\right)^{- \frac{1}{\xi}} & (\xi \neq 0), \\
      1 - \mathrm{e}^{- \frac{w - \mu}{\sigma}} & (\xi = 0)
      \end{array}
    \right. \label{eq-GPD}
\end{align}
where $\mu \in \mathbb{R}$, $\sigma > 0$, and $\xi \in \mathbb{R}$ are location, scale, and shape parameters. 

We fit these GPD parameters to the $M+1$ largest IPW weight values.
Here $M$ ($0 < M < n$) is given by heuristics;
following \citet{vehtari2024pareto}, we determine it by
\begin{align}
    M = \min \left\{ \left\lfloor \frac{n}{5} \right\rfloor, \lfloor 3 \sqrt{n} \rfloor \right\} ,
\end{align}
where $\lfloor n \rfloor$ denotes a floor function, which returns the greatest integer that is less than or equal to $n$.
Letting $w_{[1]} \leq \dots \leq w_{[n]}$
be the weights sorted in ascending order,
the $M+1$ largest ones are denoted by $w_{[n-M]}, \dots, w_{[n]}$.

Following \citet{vehtari2024pareto},
we set location parameter $\mu$ to the $(M+1)$-th largest IPW weight value, i.e.,
\begin{align}
    \hat{\mu} = w_{[n-M]}. \label{eq-mu}
\end{align}
By contrast, we estimate $\sigma$ and $\xi$ using $w_{[n-M+1]}, \dots, w_{[n]}$. 
Among several estimators, we employ the standard method termed the probability weighted moment (PWM)
\citep{hosking1987parameter},
\footnote{Although we empirically observed that 
using PWM leads to good CATE estimation performance, 
GPD parameter fitting might not be easy in general.
However, according to \citet[Section 6]{vehtari2024pareto},
one can evaluate the reliability of GPD fitting 
by employing the estimated value of GPD's shape parameter, $\hat{\xi}$,
which determines the heaviness of the distribution tail.}
which constructs the estimators of $\sigma$ and $\xi$ 
using the following weighted moment statistic:
\begin{align}
    \alpha_s = \E\left[(1 - \F(W))^s (W - \mu)\right] \quad s \in \bigl\{0, 1\bigr\}. \label{eq-alpha}
\end{align}
Roughly speaking, statistic $\alpha_s$ in \eqref{eq-alpha} is 
a weighted average of $W - \mu$ with weight $(1 - \F(W))^s$ 
and is estimated as
\begin{align}
    \hat{\alpha}_0 &= \frac{1}{M} \sum_{i=n-M+1}^n \left(w_{[i]} - \hat{\mu}\right), \label{eq-PWM-0}\\
    \hat{\alpha}_1 &= \frac{1}{M} \sum_{i=n-M+1}^n (n - i) \left(w_{[i]} - \hat{\mu}\right).  \label{eq-PWM-1}
\end{align}
Using $\hat{\alpha}_0$ and $\hat{\alpha}_1$, the PWM method estimates $\sigma$ and $\xi$ as
\begin{align}
    \hat{\sigma} &= \frac{2 \hat{\alpha}_0 \hat{\alpha}_1}{\hat{\alpha}_0 - 2 \hat{\alpha}_1}, \label{eq-sigma}\\
    \hat{\xi} &= 2 - \frac{\hat{\alpha}_0}{\hat{\alpha}_0 - 2\hat{\alpha}_1}. \label{eq-xi}
\end{align}

\subsubsection{Weight Replacement with GPD Quantiles}

Second, we replace the $M$ largest weight values, 
i.e., $w_{[n-M+1]}, \dots, w_{[n]}$,
with the quantiles of the fitted GPD with parameters $(\hat{\mu}, \hat{\sigma}, \hat{\xi})$ in \eqref{eq-mu}, \eqref{eq-sigma}, and \eqref{eq-xi}.

Since the quantile function is given by
the inverse of the cumulative distribution function,
we replace weight value $w_{[n-M+m]}$ ($m = 1, \dots, M$) 
with $\frac{m - 1/2}{M}$-quantile as
\begin{align}
    w_{[n-M+m]} = \hat{\F}^{-1}\left( \frac{m - 1/2}{M}  \right),
\end{align}
where $\hat{\F}$ denotes a fitted GPD cumulative distribution function.
By contrast, we do not change 
the other weight values, i.e., $w_{[1]}, \dots, w_{[n-M]}$.
Hence, letting $i = n - M + m$,
we can summarize the weight replacement procedure as
\begin{align}
    w_{[i]} &= \I(i \geq n-M+1)\ \hat{\F}^{-1}\left( \frac{i - (n-M) - 1/2}{M}  \right) \nonumber\\
    &+ (1 - \I(i \geq n-M+1))\ w_{[i]} \label{eq-PS}.
\end{align}

In this paper, we utilize the weight replacement formula 
in \eqref{eq-PS} 
to improve the estimation stability of weighted representation learning.
Unfortunately, we cannot directly employ this formula in an end-to-end manner
because it needs non-differentiable computations.

\subsection{Non-Differentiable Procedures} \label{sec-difficulty}

The main difficulty of using Pareto smoothing 
for weighted representation learning
is that it requires the computation of 
the \textit{rank} of each IPW weight.

Ranking is an operation that
takes input vector $\biw = [w_1, \dots, w_n]^{\top}$
and outputs the position of each element $w_i$
in sorted vector $[w_{[1]}, \dots, w_{[n]}]^{\top}$,
where $w_{[1]}\leq \dots \leq w_{[n]}$.
To illustrate this operation, consider a case with $n=3$.
For instance, 
if $\biw$ 
satisfies $w_3 \leq w_1 \leq w_2$,
since $w_1 = w_{[2]}$, $w_2 = w_{[3]}$, and $w_3 = w_{[1]}$ hold,
the rank of $\biw$ is given 
as vector $\bir = [2, 3, 1]^{\top}$.
Formally, such an operation can be expressed as 
$\bir = r(\biw)$,
using function $r$, called a rank function
(See \Cref{sec-ranking} for the definition of function $r$).

\begin{figure}[t]
    \centering
    \includegraphics[width=0.355\textwidth]{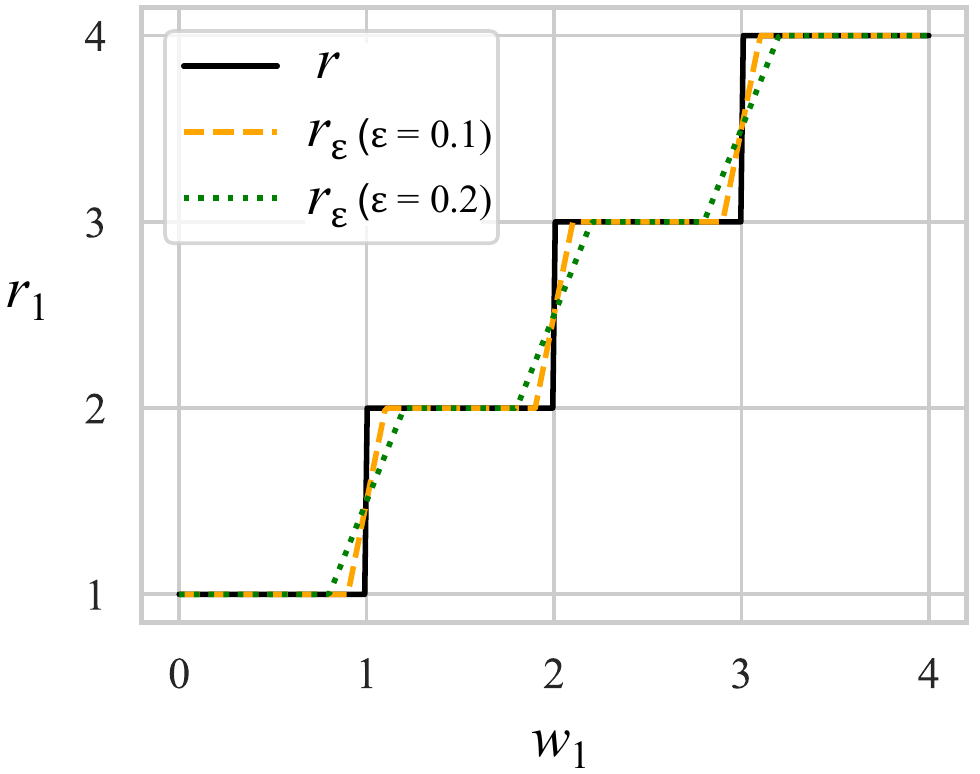}
    \caption{
        Illustration of rank function $\bir = r(\biw)$ (black) and differentiable rank functions $\bir = r_{\varepsilon}(\biw)$ (orange and green): Here we take input vector $\biw = [w_1, 1, 2, 3]^{\top}$, vary $w_1$'s value and look at how its rank $r_1 \in \bir$ changes. When regularization parameter $\varepsilon \rightarrow 0$, $r_{\varepsilon}$ converges to $r$ \citep{blondel2020fast}.}
    \label{fig2}
\end{figure}

Unfortunately, this rank function is not differentiable 
with respect to input $\biw$.
To see this, consider $\biw = [w_1, 1, 2, 3]^{\top}$
and observe how the rank of $w_1$ varies when we increase its value.
In this case,
its rank, $r_1$, is given as a piecewise constant function, 
as illustrated as the black line in \Cref{fig2}.
Since the derivative of such a piecewise constant function is 
\textbf{always} zero or undefined,
we cannot perform gradient backpropagation 
and hence cannot employ 
the weight correction technique in \eqref{eq-PS} in an end-to-end manner.
Therefore,
with such a non-differentiable rank function,
we cannot use Pareto smoothing 
for weighted representation learning,
which jointly learns the propensity score model, 
the feature representations, and the outcome prediction models.

One may consider a separate learning approach
that trains the propensity score model,
computes the Pareto-smoothed IPW weights by \eqref{eq-PS}, 
and learns the feature representations.
This approach, however,
requires directly fitting a propensity score model 
to features $\biX$, not their representations.
Since accurately estimating a propensity score model 
from high-dimensional features $\biX$ is considerably difficult,
such a separate learning approach yields
large model misspecification error
and hence leads to CATE estimation bias.
We experimentally show its poor performance in \Cref{sec-semiexp}.

For this reason,
we develop a joint learning approach 
by making the non-differentiable computation in Pareto smoothing differentiable.

\subsection{Making Pareto Smoothing Differentiable}

\subsubsection{Differentiable Approximation}

The weight replacement formula in \eqref{eq-PS} requires 
the computation of two troublesome piecewise constant functions.
One is 
rank function $r$, which is needed to obtain the position of weight $w_i$ in sorted vector $[w_{[1]}, \dots, w_{[n]}]^{\top}$,
and the other is indicator function $\I(i \geq n - M + 1)$.

To make rank function $r$ differentiable, 
we utilize the differentiable ranking technique
\citep{cuturi2019differentiable,blondel2020fast},
which approximates rank function $r(\biw)$ 
with a differentiable function.
Among the recent methods,
we select a computationally efficient one \citep{blondel2020fast},
which works with $O(n \mathrm{log}n)$ time and $O(n)$ memory complexity.
With this method, we approximate rank function $r(\biw)$
as the solution to 
the regularized linear programming (LP) 
that contains the $l^2$ regularization term 
with regularization parameter $\varepsilon > 0$.
The solution, $r_{\varepsilon}(\biw)$, is a piecewise linear function that
can well approximate rank function $r$ (as illustrated in \Cref{fig2})
and is differentiable almost everywhere,
thus greatly facilitating gradient backpropagation.

As a differentiable approximation 
of indicator function $\I$ in \eqref{eq-PS},
we employ sigmoid function $\varsigma$:
\begin{align}
    \I(i \geq j) \simeq \varsigma(i, j) \coloneqq \frac{1}{1 + \mathrm{e}^{-\kappa (i - j)}}, \label{eq-sigmoid}
\end{align}
where $\kappa > 0$ is a hyperparameter. 
 
\subsubsection{Reformulation of GPD Parameter Estimators}

To employ differentiable rank $\bir = r_{\varepsilon}(\biw)$
for Pareto smoothing,
since it represents ranks as continuous values,
we need to modify the GPD parameter estimators,
i.e., $\hat{\mu}$, $\hat{\sigma}$, and $\hat{\xi}$.

Regarding location parameter $\hat{\mu}$ in \eqref{eq-mu},
since this estimator is given as $w_{[n-M]}$, i.e., the largest weight among $w_{[1]}, \dots, w_{[n-M]}$,
we reformulate it as
\begin{align*}
    \tilde{\mu} = w_i, \ 
    \mbox{where}\ i = \arg \max_i \bigl\{r_i \mid r_i \leq n - M \bigr\}.
\end{align*}
To reformulate $\hat{\sigma}$ and $\hat{\xi}$ in \eqref{eq-sigma} and \eqref{eq-xi},
we rephrase estimators
$\hat{\alpha}_0$ and $\hat{\alpha}_1$ in \eqref{eq-PWM-0} and \eqref{eq-PWM-1}.
With non-differentiable rank $\bir = r(\biw)$,
these estimators are equivalently reformulated
by rewriting the summation over  $w_{[n-M+1]}, \dots, w_{[n]}$ 
in \eqref{eq-PWM-0} and \eqref{eq-PWM-1} with indicator function $\I$
as
\begin{align*}
    \hat{\alpha}_0 &= \frac{1}{M} \sum_{i=1}^n \I(r_i \geq n-M+1)\left(w_i - \hat{\mu}\right)  \\
    \hat{\alpha}_1 &= \frac{1}{M} \sum_{i=1}^n \I(r_i \geq n-M+1)\
    (n - r_i ) \left(w_{i} - \hat{\mu}\right).
\end{align*}
Hence, when given differentiable rank $\bir = r_{\varepsilon}(\biw)$,
by replacing indicator function $\I$ with sigmoid function $\varsigma$ 
in \eqref{eq-sigmoid},
we make $\hat{\alpha}_0$ and $\hat{\alpha}_1$ 
differentiable with respect to $\bir$:
\begin{align}
    \tilde{\alpha}_0 &= \frac{1}{\tilde{M}} \sum_{i=1}^n \varsigma(r_i, n-M+1)\left(w_i - \tilde{\mu}\right), \label{eq-PWM-0d}  \\
    \tilde{\alpha}_1 &= \frac{1}{\tilde{M}} \sum_{i=1}^n \varsigma(r_i, n-M+1)\
    (n - r_i) \left(w_{i} - \tilde{\mu}\right),  \label{eq-PWM-1d} 
\end{align}
where $\tilde{M} = \sum_{i=1}^n \varsigma(r_i, n-M+1)$. 
By substituting $\tilde{\alpha}_0$ and $\tilde{\alpha}_1$ 
for $\hat{\alpha}_0$ and $\hat{\alpha}_1$ in \eqref{eq-sigma} and \eqref{eq-xi}, we compute scale and shape parameters as 
$\tilde{\sigma}$ and $\tilde{\xi}$, respectively.

\subsubsection{Overall Algorithm}

Using the GPD cumulative distribution function, $\tilde{\F}$, 
with parameters $(\tilde{\mu},  \tilde{\sigma}, \tilde{\xi})$, 
we replace each weight $w_i$ in \eqref{eq-IPW} with
\begin{align}
    \tilde{w}_i &= \varsigma(r_i, n-M+1)\ \tilde{\F}^{-1}\left( \zeta \left(
        \frac{r_i - (n-M) - 1/2}{M}
        \right) \right) \nonumber\\
    &+ (1 - \varsigma(r_i, n-M+1))\ w_i, \label{eq-dif-replace-w}
\end{align}
where $\zeta(x) \coloneqq \min \left\{\max \left\{x, 0\right\}, 1\right\}$ 
is a function that forces input $x$ to lie in $[0, 1]$. 
Using Pareto-smoothed weight $\tilde{w}_i$ instead of $w_i$,
we minimize the objective function in \eqref{eq-obj-other}.

\Cref{alg1} summarizes our method.
To alternately minimize the objective functions 
in \eqref{eq-obj-prop} and \eqref{eq-obj-other},
we perform stochastic gradient descent \citep{kingma2015}.
After the convergence, 
we estimate the CATE in \eqref{eq-CATE}
by $h^1(\Delta(\bix), \Upsilon(\bix)) - h^0(\Delta(\bix), \Upsilon(\bix))$.

Compared with the DRCFR method \citep{hassanpour2020learning}, 
our method requires additional time 
to compute Pareto-smoothed weights (lines 12-16 in \Cref{alg1}).
In particular, computing differentiable rank (line 12) 
requires time complexity $\mathcal{O}(B \mathrm{log} B)$ 
for mini-batch size $B$,
which is needed to evaluate the objective function 
in \eqref{eq-obj-other} and its gradient 
for each iteration in the training phase. 

\textbf{Remark}: 
Strictly speaking, 
the choice of activation functions in propensity score $\pi$ and 
feature representations $\Gamma$ and $\Delta$
is critical for satisfying the assumption of Pareto smoothing
that the distribution of the importance sampling weight 
is absolutely continuous,
which is necessary to prove the asymptotic consistency 
(Theorem 1 of \citet{vehtari2024pareto}). 
This assumption holds
if each activation is differentiable almost everywhere 
(i.e., differentiable except on a set of measure zero).
However, for instance, 
using the rectified linear unit (ReLU) 
in propensity score model $\pi$
makes the distribution of IPW weight discontinuous, 
thus violating the assumption of Pareto smoothing.
Even with almost everywhere differentiable activation functions,
due to the lack of learning theory on neural network models,
deriving the asymptotic consistency of our CATE estimator 
is extremely challenging
and is left as our future work.

\begin{algorithm}[t]
    \caption{Differentiable Pareto-Smoothed Weighting (DPSW)} \label{alg1}
    \begin{algorithmic}[1]
    \STATE Initialize the parameters of $\Gamma$, $\Delta$, $\Upsilon$, $\pi$, $h^0$, and $h^1$
    \WHILE {not converged}
    \WHILE {not converged}
    \STATE Sample mini-batch from $\mathcal{D} = \{(a_i, \bix_i, y_i)\}_{i=1}^n$
    \STATE Update $\pi$ by minimizing cross entropy loss in \eqref{eq-obj-prop}
    \ENDWHILE
    \WHILE {not converged}
    \STATE Sample mini-batch from $\mathcal{D} = \{(a_i, \bix_i, y_i)\}_{i=1}^n$
        \FOR{instance $i$ in mini-batch}
        \STATE Compute weight $w_i$ by \eqref{eq-IPW}
        \ENDFOR
        \STATE Compute differentiable rank $\bir = r_{\varepsilon}(\biw)$
        \STATE Estimate GPD parameters as $\tilde{\mu}$, $\tilde{\sigma}$, and $\tilde{\xi}$
        \FOR{instance $i$ in mini-batch}
        \STATE Replace each weight $w_i$ with $\tilde{w}_i$ in \eqref{eq-dif-replace-w}
        \ENDFOR
        \STATE Update $\Gamma$, $\Delta$, $\Upsilon$, $h^0$, and $h^1$ by minimizing prediction loss in $\eqref{eq-obj-other}$ with Pareto-smoothed weights $\{\tilde{w}_i\}$
    \ENDWHILE
    \ENDWHILE
    \end{algorithmic} 
\end{algorithm}

\section{Experiments} \label{sec-exp}

\subsection{Semi-synthetic Data} \label{sec-semiexp}

First, we evaluated the CATE estimation performance 
using semi-synthetic benchmark datasets, 
where the true CATE values are available, unlike real-world data.

\textbf{Data:} We selected the two high-dimensional datasets: 
the News 
and the Atlantic Causal Inference Conference (ACIC) datasets 
\citep{johansson2016learning,shimoni2018benchmarking}. 

The News dataset is constructed from $n=5000$ articles,
randomly sampled from the New York Times corpus 
in the UCI repository.
\footnote{\url{https://archive.ics.uci.edu/dataset/164/bag+of+words}}
The task is to infer the effect of 
the viewing device 
(desktop ($A=0$) or mobile ($A=1$)) 
on the readers' experience $Y$. 
Features $\biX$ are the count of $d=3477$ words in each article.
Treatment $A$ and outcome $Y$ are simulated
using the latent topic variables obtained 
by fitting a topic model on $\biX$.
The ACIC dataset is derived from the clinical measurements of $d=177$ features in the Linked Birth and Infant Death Data (LBIDD) \citep{macdorman1998infant},
which was developed for a data analysis competition called ACIC2018.
We randomly selected $n=5000$ observations and prepared $20$ datasets.
With both semi-synthetic datasets, 
we randomly split each sample 
into training, validation, and test data 
with a ratio of 60/20/20.

\textbf{Baselines:} To evaluate
our method (\textbf{DPSW}) 
and its variant that performs self-normalization (\textbf{DPSW Norm.}),
we consider $15$ baselines.
With DRCFR \citep{hassanpour2020learning},
we tested four different weighting schemes:
no weight modification (\textbf{DRCFR}),
self-normalization (\textbf{DRCFR Norm.}; Eq. \eqref{eq-Norm}),
 weight truncation with a threshold suggested by \citet{crump2009dealing} (\textbf{DRCFR Trunc.}; Eq. \eqref{eq-Trunc}),
and a scheme that ignores the prediction loss for individuals with extreme weights based on the same threshold (\textbf{DRCFR Ignore}).
We also tested a separate learning approach (\textbf{PSW}; \Cref{sec-difficulty}), 
which trains propensity score $\pi$ with $\{(a_i, \bix_i)\}_{i=1}^n$ beforehand
and learns only $\Delta(\biX)$ and $\Upsilon(\biX)$
using Pareto-smoothed IPW weights.
Other baselines include
(i) linear regression methods:
a single model with treatment $A$ added to its input (\textbf{LR-1})
and two separate models for each treatment (\textbf{LR-2});
(ii) meta-learner methods: 
the S-Learner (\textbf{SL}),
the T-Learner (\textbf{TL}),
the X-Learner (\textbf{XL}),
and the DR-Learner (\textbf{DRL});
(iii) tree-based methods:
causal forest \citep{athey2019generalized} (\textbf{CF})
and a variant combined with double/debiased machine learning 
\citep{chernozhukov2018double} (\textbf{CF DML});
and (iv) neural network methods:
the treatment-agnostic regression network \citep{shalit2017estimating} (\textbf{TARNet})
and the generative adversarial network \citep{yoon2018ganite} (\textbf{GANITE}).
We describe the settings of these baselines in \Cref{sec-impl}.

\textbf{Settings:} Regarding our method and DRCFR, 
we used three-layered feed-forward neural networks (FNNs)
to formulate feature representations 
$\Gamma(\biX)$, $\Delta(\biX)$, and $\Upsilon(\biX)$,
propensity score $\pi$,
and outcome prediction models 
$h^0$ and $h^1$.

We tuned the hyperparameters 
(e.g., parameter $\varepsilon$ 
of differentiable rank $r_{\varepsilon}(\biw)$ in our method)
by minimizing the objective function value on the validation data;
such hyperparameter tuning is standard 
for CATE estimation \citep{shalit2017estimating}.

\textbf{Performance metric:} Following \citet{hill2011bayesian}, we used the precision in the estimation of heterogeneous effect (PEHE), $\mathrm{PEHE} \coloneqq \sqrt{\frac{1}{n} \left((y^1_i - y^0_i) - \hat{\tau}_i \right)^2}$, where $y^0_i$ and $y^1_i$ are the true potential outcomes, and $\hat{\tau}_i$ denotes the predicted CATE value. We computed the mean and the standard deviation of the test PEHEs over $50$ realizations of potential outcomes (the News dataset) and $20$ realizations (the ACIC dataset).

\textbf{Results:} \Cref{table-PEHE} presents the test PEHEs on the News and ACIC datasets.

\begin{table}[t]
    \caption{Mean and standard deviation of test PEHE on semi-synthetic datasets (lower is better)}
    \centering
    \label{table-PEHE}
    \begin{tabular}{lcc}
    \toprule
    Method & News ($d=3477$) & ACIC ($d=177$) \\
    \midrule
    LR-1          & $3.35 \pm 1.28$ & $0.72 \pm 0.07$ \\
    LR-2          & $5.36 \pm 1.75$ & $3.82 \pm 0.15$     \\
    \midrule
    SL           & $2.83 \pm 1.11$  & $1.69 \pm 0.52$     \\
    TL           & $2.55 \pm 0.82$  & $2.23 \pm 0.50$     \\
    XL           & $2.77 \pm 1.01$  & $1.05 \pm 0.72$     \\
    DRL          & $23.9 \pm 5.96$  & $3.77 \pm 8.96$     \\
    \midrule
    CF            & $3.84 \pm 1.67$ & $3.55 \pm 0.19$   \\
    CF DML        & $2.69 \pm 1.06$ & $1.18 \pm 0.32$ \\
    \midrule
    TARNet        & $4.92 \pm 1.80$ & $3.31 \pm 0.51$     \\
    GANITE        & $2.68 \pm 0.66$ & $3.69 \pm 0.77$     \\
    \midrule
    DRCFR        & $2.38 \pm 0.66$ & $0.98 \pm 0.07$     \\
    DRCFR Norm.  & $2.37 \pm 0.94$ & $0.73 \pm 0.12$     \\
    DRCFR Trunc. & $2.42 \pm 0.79$ & $1.06 \pm 0.06$     \\
    DRCFR Ignore & $2.35 \pm 0.75$ & $0.84 \pm 0.06$     \\
    PSW           & $4.03 \pm 1.35$ & $0.71 \pm 0.01$     \\
    \midrule
    \textbf{DPSW}  & \textbf{2.20} $\pm$ \textbf{0.72} &  \textbf{0.57} $\pm$ \textbf{0.03}    \\
    \textbf{DPSW Norm.} & \textbf{2.10} $\pm$ \textbf{0.66}      & \textbf{0.52} $\pm$ \textbf{0.16}    \\
    \bottomrule
    \end{tabular}
\end{table}

Our proposed frameworks (\textbf{DPSW} and \textbf{DPSW Norm.}) outperformed all the baselines, 
demonstrating their effectiveness 
in CATE estimation from high-dimensional data.
\textbf{DPSW Norm.} achieved lower PEHEs than \textbf{DPSW},
implying that the self-normalization of Pareto-smoothed weights further 
improves the stability of the weight estimation.

Weighted representation learning methods (DR-CFR and DPSW) outperformed the other neural network methods (TARNet and GANITE),
especially on the ACIC dataset. 
Given that treatment $A$ and outcome $Y$ of this dataset were simulated 
using different features in $\biX$,
these results emphasize the importance of
performing data-driven feature separation 
by weighted representation in such a setup.

PSW worked much worse on the News dataset than DPSW,
indicating that fitting a propensity score
directly to high-dimensional features $\biX$ 
leads to severe model misspecification error,
complicating subsequent weighted representation learning, 
even with Pareto-smoothed weights.
By contrast, our joint learning approach performed well,
since it used differentiable Pareto smoothing.

\subsection{Synthetic Data} \label{sec-synthexp}

Next we investigated 
how well our method learned the feature representations
using synthetic data,
where the data-generating processes are entirely known.

\begin{figure*}[t]
    \centering
    \includegraphics[width=0.975\textwidth]{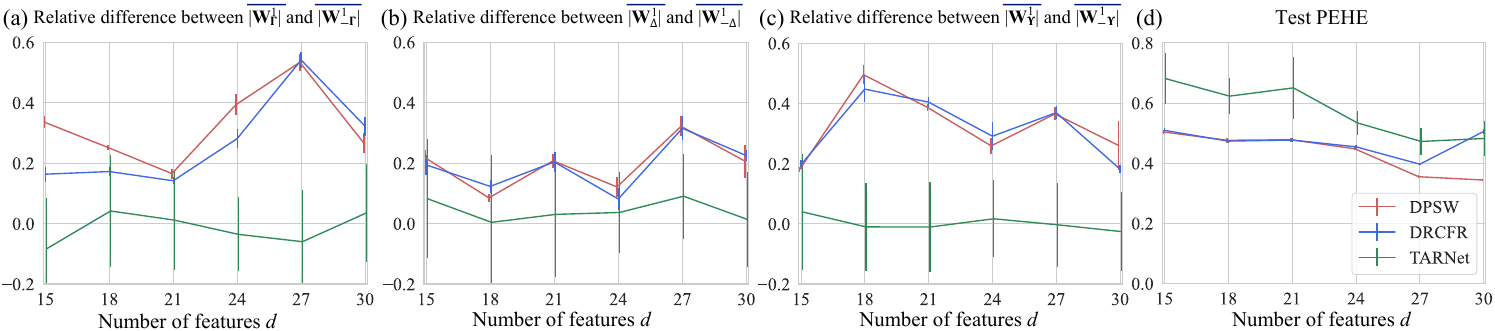}
    \caption{Learned encoder parameter differences and test PEHEs on synthetic data: 
    (a): value difference of $\mathbf{W}^1$ in encoder $\Gamma(\biX)$; 
    (b): value difference of $\mathbf{W}^1$ in encoder $\Delta(\biX)$; 
    (c): value difference of $\mathbf{W}^1$ in encoder $\Upsilon(\biX)$; 
    (d) test PEHEs.
    With TARNet, since it learns a single encoder,
    we computed all parameter value differences 
    with weight matrix in same encoder.}
    \label{fig3}
\end{figure*}

\textbf{Data:} Following \citet{hassanpour2020learning}, 
we simulated the synthetic data.
We randomly generated features 
$\biX = [\biX_{\Gamma}, \biX_{\Delta}, \biX_{\Upsilon}]^{\top} \in \R^{d}$ 
($d = 15, 18, \dots, 30$).
Next, by regarding feature subsets $\biX_{\Gamma} \in \R^{d/3}$, 
$\biX_{\Delta} \in \R^{d/3}$, 
and $\biX_{\Upsilon} \in \R^{d/3}$ as instrumental variables, 
confounders, and adjustment variables, respectively,
we sampled binary treatment $A$ 
using $\biX_{\Gamma}$ and $\biX_{\Delta}$ 
and outcome $Y$ by employing $\biX_{\Delta}$ and $\biX_{\Upsilon}$
(See \Cref{sec-synth-data} for details).
We split each of the $20$ datasets ($n=20000$)
with a 50/25/25 training/validation/test ratio.

\textbf{Performance metric:} As with \citet{hassanpour2020learning},
we evaluated the quality of the learned feature representations 
$\Gamma(\biX)$, $\Delta(\biX)$, and $\Upsilon(\biX)$,
each of which is formulated as a three-layered FNN encoder:
\begin{align*}
    \mathrm{FNN}(\biX) \coloneqq \nu\left(\biW^3 \nu\left(\biW^2 \nu\left(\biW^1 \biX\right)\right)\right), 
\end{align*}
where $\nu$ is the exponential linear units (ELUs) \citep{clevert2016accurate}, 
and $\biW^1$, $\biW^2$, and $\biW^3$ are the weight parameter matrices in the first, second, and third layer, respectively.

To determine whether the learned FNN encoders 
correctly look at important features,
we measured the attribution of features 
$\biX_{\Gamma}$, $\biX_{\Delta}$, and $\biX_{\Upsilon}$
by employing $\biW^1$, i.e., the trained weight matrix 
in the first layer of each encoder.
For instance, we quantified $\biX_{\Gamma}$'s attribution on $\Gamma(\biX)$
in two steps.
First, we partitioned its learned weight parameter matrix as
$\biW^1 = [\biW^1_{\Gamma}, \biW^1_{-\Gamma}]$,
where $\biW^1_{\Gamma}$ is a submatrix with the first $d/3$ columns of $\biW^1$
and $\biW^1_{-\Gamma}$ is the one with the other columns.
Then we measured how greatly features $\biX_{\Gamma}$ affect
learned representation $\Gamma(\biX)$
by taking the relative difference between the average absolute values of the weight parameter submatrices, i.e.,
    $\frac{\overline{| \biW^1_{\Gamma}|} - \overline{|\biW^1_{-\Gamma}|}}{\overline{|\biW^1_{-\Gamma}|}}$.
We evaluated other learned representations, $\Delta(\biX)$ and $\Upsilon(\biX)$, in the same way.

\textbf{Results:} 
\Cref{fig3} shows the means and standard deviations 
of the learned parameter differences and the test PEHEs
over $20$ randomly generated synthetic datasets.

With our DPSW method and DRCFR,
absolute parameter values 
$|\biW^1_{\Gamma}|$, $|\biW^1_{\Delta}|$, and $|\biW^1_{\Gamma}|$
were sufficiently larger than 
$|\biW^1_{-\Gamma}|$, $|\biW^1_{-\Delta}|$, and $|\biW^1_{-\Gamma}|$, respectively,
showing that 
both methods correctly learned 
$\Gamma(\biX)$, $\Delta(\biX)$, and $\Upsilon(\biX)$
that are highly dependent on
instrumental variables $\biX_{\Gamma}$, 
confounders $\biX_{\Delta}$, 
and adjustment variables $\biX_{\Upsilon}$, respectively.
These results offer a clear contrast to TARNet, 
which learns a single representation
without feature separation.

The same is true for the CATE estimation performance (\Cref{fig3} (d)).
TARNet's test PEHE was larger than DPSW and DRCFR,
demonstrating the importance of data-driven feature separation 
by weighted representation learning.
By contrast, our method achieved the lowest PEHE,
thus indicating that our weight correction framework 
successfully improved the CATE estimation performance of DRCFR.

\textbf{Performance under high-dimensional setup:}
We confirmed that our method also worked well with $d=600, 1200, \dots, 3000$
(See \Cref{sec-add-exp} for details).

\section{Related Work}

\textbf{Data-driven feature separation for CATE estimation:}
CATE estimation has gained increasing attention
because of its great importance 
for causal mechanism understanding \citep{chikahara22a,zhao2022selective} and for decision support in various fields,
such as precision medicine \citep{gao2021assessment} 
and online advertising \citep{sun2015causal}.
There has been a surge of interest 
in leveraging flexible machine learning models,
including tree-based models \citep{athey2019generalized,hill2011bayesian},
Gaussian processes \citep{alaa2018bayesian,horii2024uncertainty},
and neural networks \citep{hassanpour2019counterfactual,johansson2016learning,shalit2017estimating}.
However, 
most methods treat all input features $\biX$ as confounders.
As pointed out by \citet{wu2022learning},
the empirical performance of such methods greatly varies 
with the presence of 
adjustment variables in $\biX$,
which is usual in practice, especially in high-dimensional settings.

Motivated by this issue,
we develop data-driven feature separation methods 
for treatment effect estimation. 
A pioneering work is data-driven variable decomposition (D$^2$VD) 
\citep{kuang2017treatment,kuang2020data},
which minimizes the weighted prediction loss 
plus the regularizer for feature separation.
A recent method addresses a more complicated setup,
where features $\biX$ include \textit{post-treatment variables},
which are affected by treatment $A$ 
\citep{wang2023treatment}.
However, the estimation target of these methods is ATE,
not CATE.

By contrast,
DRCFR deals with CATE estimation and 
is founded on weighted representation learning,
which is a promising approach for addressing high-dimensional data.
These advantages are why we adopted it
as the inference engine of our weight correction framework.
Integrating the recent idea of enforcing independence 
between feature representations with mutual information \citep{cheng2022learning,chu2022disentangled,liu2024edvae} 
remains our future work.

\textbf{Weighting schemes for treatment effect estimation:}
IPW \citep{rosenbaum1983central} is a common weighting technique 
for treatment effect estimation.
However, a weighted estimator based on IPW 
is often numerically unstable
due to the computation of the inverse of propensity scores.
One remedy 
is weight truncation \citep{crump2009dealing},
which, however, causes estimation bias, 
leading to inaccurate treatment effect estimation.

To improve the estimation performance,
\citet{zhu2020pareto} employed Pareto smoothing \citep{vehtari2024pareto}.
Although they empirically show that
using this technique leads to better performance than weight truncation, 
their method was developed to estimate ATE, not CATE.

Applying Pareto smoothing in weighted representation learning for CATE estimation is difficult 
because it prevents gradient backpropagation due to non-differentiability. 
This difficulty is disappointing, 
given that previous work has theoretically shown that
such weight correction schemes as weight truncation 
help extract predictive feature representations 
for CATE estimation \citep{assaad2021counterfactual}.

To establish a Pareto-smoothed weighting framework 
for CATE estimation from high-dimensional data,
we demonstrated how 
a differentiable ranking technique \citep{blondel2020fast} 
can be used to simultaneously learn a propensity score model
and feature representations. 

\section{Conclusion}

We established a differentiable Pareto-smoothed weighting framework 
for CATE estimation from high-dimensional data.
To construct a CATE estimator 
that is numerically robust to propensity score estimation error,
we develop a differentiable weight correction procedure 
based on Pareto smoothing 
and incorporated it into 
weighted representation learning for CATE estimation.
We experimentally show that
our framework outperformed 
traditional weighting schemes
as well as the existing CATE estimation methods.

By leveraging the versatility of weighting,
our future work will investigate how to extend our framework 
to estimate the effects of 
high-dimensional binary treatment \citep{zou2020counterfactual},
continuous-valued treatment \citep{wang2022generalization},
and time series treatment \citep{lim2018forecasting}.

\bibliography{main}

\clearpage
\appendix
\setcounter{theorem}{0}
\setcounter{proposition}{0}

\title{Supplementary Materials for\\ 
"Differentiable Pareto-Smoothed Weighting \\
for High-Dimensional Heterogeneous Treatment Effect Estimation"}

\onecolumn

\maketitle

\section{Rank Function Definition} \label{sec-ranking}

In \Cref{sec-difficulty},
we consider rank function $r$,
which takes input vector $\biw = [w_1, \dots, w_n]^{\top}$
and outputs the rank of each element in $\biw$.
We formally define this rank function
based on a sorting operation and the concept of inverse permutation.

Consider a sorting operation over $w_1, \dots, w_n$ in $\biw \in \R^n$ 
that finds permutation $\birho = [\rho_1, \dots, \rho_n]^{\top}$
such that the vector values that are permuted according to $\birho$,
$\biw_{\birho} = [w_{\rho_1}, \dots, w_{\rho_n}]^{\top}$,
are increasing
as $w_{\rho_1} \leq \dots \leq w_{\rho_n}$.
Let $\birho^{-1}$ be the inverse of permutation $\birho$,
i.e., a permutation whose $\rho_i$-th element 
is $i$ for $i = 1, \dots, n$.

Then the ranking function is defined as the inverse of the sorting permutation:
\begin{align} 
    r(\biw) = \birho^{-1}(\biw).
\end{align}
Throughout this paper, we consider rank function $r(\biw)$ 
that evaluates the position of each $w_i$ 
based on sorting in an ascending order. 
If we need to address the ranking in descending order, 
we can formulate it as $r(-\biw)$.

\section{Experimental Details} \label{sec-exp-setting}

\subsection{Settings of Baselines} \label{sec-impl}

Regarding linear regression methods (LR-1 and LR-2),
to avoid overfitting due to the large number of input features,
we employed the ridge regression model
in scikit-learn \citep{pedregosa2011scikit}.
For the meta-learner methods
(i.e., the S-Learner (SL), the T-Learner (TL), the X-Learner (XL), 
the DR-Learner (DRL)) and the tree-based methods (CF and CF DML)
we used the EconML Python package \citep{battocchi2019econml}.
As the base learners for SL, TL, XL, and DRL, 
we chose random forest
because we empirically observed that 
it achieved the best performance among the three model candidates:
random forest, gradient boosting, and support vector machine.

We evaluated the performance of the neural network methods 
(TARNet and GANITE), 
with the existing implementations 
\citep{curth2021nonparametric,yoon2018ganite}.
\footnote{\url{https://github.com/AliciaCurth/CATENets}}
\textsuperscript{,}
\footnote{\url{https://github.com/jsyoon0823/GANITE}}
By contrast, we implemented the DRCFR method
using PyTorch \citep{paszke2019pytorch}.

With PSW, we trained propensity score model $\pi(\biX)$ beforehand,
using a paired sample \{$(a_i, \bix_i)$\} $\subset \mathcal{D}$. 
We formulated $\pi(\biX)$ using a three-layered FNN, 
as with our method and DRCFR.
After computing the IPW weights based on the trained propensity score model,
we performed Pareto smoothing over them
with the \texttt{psislw()} function 
in a Python package called ArviZ \citep{arviz_2019}.
Using the Pareto-smoothed weights,
we learned two encoders $\Delta(\biX)$ and $\Upsilon(\biX)$,
as well as outcome prediction models $h^0$ and $h^1$.

\subsection{Synthetic Data} \label{sec-synth-data}

Following \citet{hassanpour2020learning},
we prepared synthetic datasets.\footnote{We used their implementation in
\url{https://www.dropbox.com/sh/vrux2exqwc9uh7k/AAAR4tlJLScPlkmPruvbrTJQa?dl=0}.} 

We first drew the values of features $\biX \in \R^{\nf}$ from standard multivariate Gaussian distribution:
\begin{align}
    \biX \sim \mathcal{N}(0, I),
\end{align} 
where $\mathcal{N}$ denotes the Gaussian distribution.

Next, we sampled the values of treatment $A$ and outcome $Y$ 
using $\biX_{\Gamma} \in \R^{d/3}$, 
$\biX_{\Delta} \in \R^{d/3}$, 
and $\biX_{\Upsilon} \in \R^{d/3}$,
which are the feature subsets in 
$\biX = [\biX_{\Gamma}, \biX_{\Delta}, \biX_{\Upsilon}]^{\top}$.
In particular,
we employed $\psi = [\biX_{\Gamma}, \biX_{\Delta}]^{\top} \in \R^{2d/3}$
to generate $A$'s values as
\begin{align}
    A \sim \mathrm{Ber}\left(
        \frac{1}{1 + \mathrm{exp}(-\bic_A \cdot (\psi + \mathbf{1}))}
    \right),
\end{align}
where $\bic_A \in \R^{2d/3}$ is a coefficient vector 
drawn from $\mathcal{N}(0, 1)$, 
and $\mathbf{1} = [1, \dots, 1]^{\top}$ is a vector with length $2d/3$. 
By contrast, we used 
$\phi = [\biX_{\Delta}, \biX_{\Upsilon}]^{\top} \in \R^{2d/3}$
to simulate potential outcomes $\Yz$ and $\Yo$ as
\begin{align}
    \Yz &= \frac{3}{2d} \bic_{\Yz} \cdot \phi + \epsilon \\
    \Yo &= \frac{3}{2d} \bic_{\Yo} \cdot (\phi \odot \phi)  + \epsilon, \\
\end{align}
where $\bic_{\Yz} \in \R^{2d/3}$ and $\bic_{\Yo} \in \R^{2d/3}$ 
are the coefficient vectors drawn from $\mathcal{N}(0, 1)$, 
symbol $\odot$ denotes 
the element-wise product (a.k.a., Hadamard product),
and $\epsilon \sim \mathcal{N}(0, 1)$ is scalar standard Gaussian noise.
Using the values of potential outcomes $\Yz$ and $\Yo$,
we computed the values of outcome $Y$ by $Y =  A \Yo + (1-A) \Yz$.

\section{Additional Experimental Results} \label{sec-add-exp}

In \Cref{sec-synthexp}, 
as with \citet{hassanpour2020learning},
we present the performance on relatively low-dimensional synthetic data.
To further investigate its performance,
we ran additional experiments 
using high-dimensional synthetic data.

Here we generated the synthetic data in the same way 
(described in \Cref{sec-synth-data})
except that the number of features $d$ was set to 
$d = 600, 1200, \dots, 3000$.

\begin{figure*}[t]
    \centering
    \includegraphics[width=0.9\textwidth]{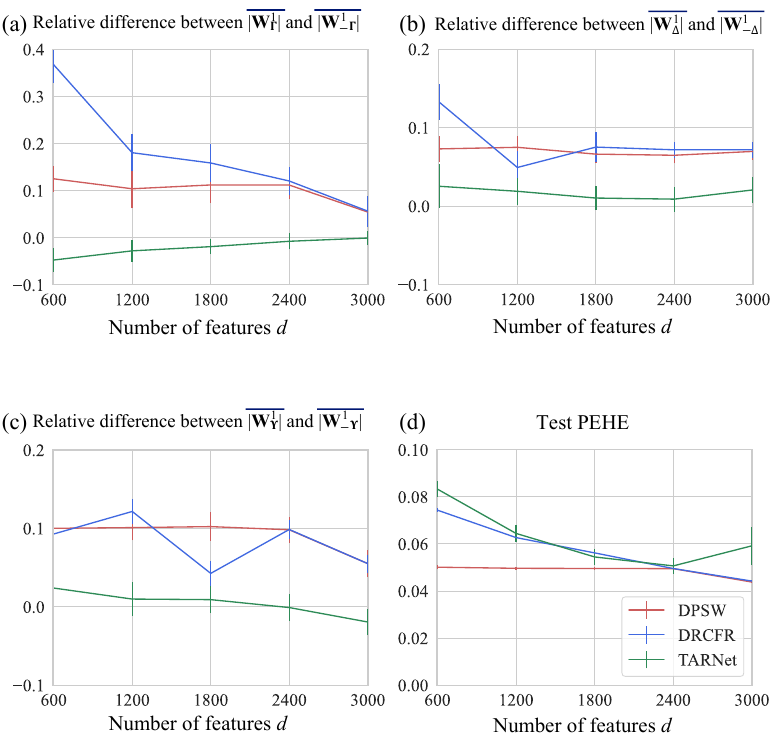}
    \caption{Learned encoder parameter differences and test PEHEs on synthetic data:
    (a): value difference of $\mathbf{W}^1$ in encoder $\Gamma(\biX)$; 
    (b): value difference of $\mathbf{W}^1$ in encoder $\Delta(\biX)$; 
    (c): value difference of $\mathbf{W}^1$ in encoder $\Upsilon(\biX)$; 
    (d) test PEHEs.
    With TARNet, since it learns single encoder,
    we computed all parameter value differences 
    with weight matrix in same encoder.}
    \label{fig-appendix}
\end{figure*}

\Cref{fig-appendix} shows the results.
Again, our method achieved better CATE estimation performance 
than DRCFR and TARNet,
thus showing that it successfully improved 
the estimation performance of DRCFR
by effectively correcting the IPW weights 
by differentiable Pareto smoothing.

Regarding the feature representation learning performance (\Cref{fig-appendix} (a) - (c)),
both DRCFR and our method outperformed TARNet.
As illustrated in \Cref{fig-appendix} (a), however,
with DRCFR and our method,
the difference among 
the absolute parameter values in learned encoder $\Gamma(\biX)$ decreased, 
as the number of features $d$ increased,
indicating that 
distinguishing the representation of 
instrumental variables $\Gamma(\biX)$ from 
that of the confounders $\Delta(\biX)$
was difficult for both methods.
This difficulty arises partly because 
binary treatment $A$ can be accurately predicted solely from
the representation of the confounders (i.e., $\Delta(\biX)$)
when there are a sufficient number of confounders $\biX_{\Delta}$.

A possible solution is to 
enforce the independence between feature representations
using an additional mutual information regularizer, 
as with \citet{cheng2022learning}.
Our future work constitutes the evaluation of 
such a variant of our method for CATE estimation performance 
under high-dimensional setups.

\end{document}